\pdfoutput=1
\documentclass{article}
\usepackage{ieeetrantools}
\usepackage{subcaption}
\usepackage{spconf,amsmath,graphicx}
\usepackage{setspace}
\usepackage{tabularx,booktabs}
\usepackage{diagbox}
\usepackage{float}
\usepackage{multirow} 
\usepackage[hyphens]{url}
\usepackage[pagebackref,breaklinks,colorlinks]{hyperref}
\usepackage{xcolor}
\usepackage{colortbl}
\usepackage{amsfonts, amsmath, amsthm, amssymb}
\usepackage{tikz}
\usepackage{textcomp}
\usepackage{lipsum}
\usepackage{svg}
\usepackage{enumitem}
\usepackage[nameinlink]{cleveref}
\usepackage{pifont}
\usepackage{scalerel}
\usepackage{stackengine}
\newcommand{\cmark}{\ding{51}}%
\newcommand{\xmark}{\ding{55}}%
\setlist[itemize]{noitemsep}
\makeatletter

\newcommand*{\overrightharpoonup}{\mathpalette{\overarrow@\rightharpoonupfill@}}
\newcommand*{\rightharpoonupfill@}{\arrowfill@\relbar\relbar\rightharpoonup}
\makeatother

\makeatletter
\newcommand*{\overleftharpoonup}{\mathpalette{\overarrow@\leftharpoonupfill@}}
\newcommand*{\leftharpoonupfill@}{\arrowfill@\relbar\relbar\leftharpoonup}
\makeatother

\definecolor{gray90}{gray}{0.9}

\makeatletter
\newcommand*\bigcdot{\mathpalette\bigcdot@{1}}
\newcommand*\bigcdot@[2]{\mathbin{\vcenter{\hbox{\scalebox{#2}{$\m@th#1\bullet$}}}}}
\makeatother
\newcommand\openbigstar[1][0.7]{%
  \scalerel*{%
    \stackinset{c}{-.125pt}{c}{}{\scalebox{#1}{\color{white}{$\bigstar$}}}{%
      $\bigstar$}%
  }{\bigstar}
}
\crefname{section}{Section}{Sections}
\crefname{equation}{Equation}{Equations}
\crefname{figure}{Figure}{Figures}
\crefname{table}{Table}{Tables}
\crefname{listing}{Listing}{Listings}

\setlist[itemize]{leftmargin=10pt,itemindent=0pt,topsep=2pt,partopsep=2.5pt,parsep=1pt,itemsep=1.5pt,listparindent=\parindent}
\setlist[enumerate]{leftmargin=16pt,itemindent=0pt,topsep=2pt,partopsep=2.5pt,parsep=1pt,itemsep=1.5pt,listparindent=\parindent}

\title{Real-time Video Prediction With Fast Video Interpolation Model \\ and Prediction Training}
\name{Shota Hirose$^1$, Kazuki Kotoyori$^1$, Kasidis Arunruangsirilert$^1$, Fangzheng Lin$^2$, Heming Sun$^{3}$, Jiro Katto$^{1}$}
\address{$^1$School of Fundamental Science and Engineering, Waseda University, Tokyo, Japan \\ 
$^2$School of Engineering, Tokyo Institute of Technology, Tokyo, Japan \\
$^3$Yokohama National University, Kanagawa, Japan }

\begin{document}
\bstctlcite{IEEEexample:BSTcontrol}
%
\maketitle

\begin{abstract}

Transmission latency significantly affects users’ quality of experience in real-time interaction and actuation. As latency is principally inevitable, video prediction can be utilized to mitigate the latency and ultimately enable zero-latency transmission. However, most of the existing video prediction methods are computationally expensive and impractical for real-time applications. In this work, we therefore propose real-time video prediction towards the zero-latency interaction over networks, called IFRVP (Intermediate Feature Refinement Video Prediction). Firstly, we propose three training methods for video prediction that extend frame interpolation models, where we utilize a simple convolution-only frame interpolation network based on IFRNet. Secondly, we introduce ELAN-based residual blocks into the prediction models to improve both inference speed and accuracy. Our evaluations show that our proposed models perform efficiently and achieve the best trade-off between prediction accuracy and computational speed among the existing video prediction methods. A demonstration movie is also provided at \textbf{http://bit.ly/IFRVPDemo}. \\
The code will be released at \href{https://github.com/FykAikawa/IFRVP}{https://github.com/FykAikawa/\\IFRVP}.

\end{abstract}
%

\section{Introduction}
\label{sec:intro}

Advancement in telecommunication technology drives implementations of video communication in many applications, such as video surveillance, virtual conferences, remote working, online gaming, real-time live streaming, etc. In these applications, reducing latency is not only highly beneficial for real-time interaction, but also especially significant in mission-critical systems such as remote surgery, remote operation with haptic devices, and self-driving vehicles: low-latency telemetry data is required for safety operations. 

\begin{figure}[t!]
\centering\includegraphics[width=\linewidth]{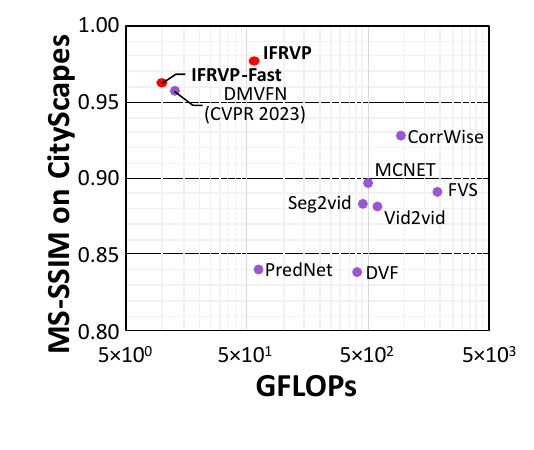}
\caption{MS-SSIM on Cityscapes \cite{cityscapes} (resized to 512$\times$1024) vs. computational complexity of various video prediction models. Proposed methods are marked as \textcolor{red}{$\bigcdot$}. We use the reported values from \cite{DMVFN} for other methods.}
\label{fig:Comparison}

\vspace{-6mm}
\end{figure}

Many factors, including network congestion and processing delays, resulted in latency. While advancements in network technology such as the 5G New Radio (NR) network, Beyond 5G (or 6G), and all-optical networks greatly reduce end-to-end latency, zero latency still remains technically impossible due to the physical speed limit of radio waves and light, and although new hardware codecs greatly reduce video processing delay, the increasing demand for higher video quality, e.g., 4K and 8K contents, generally offsets these improvements.

Currently, we are attempting to demonstrate zero-latency interactive transmission by combining low-latency network technology with deep-learning-based prediction at the application layer \cite{Teleoperation,Compensation,5GNR,Zero}. Independently, recent work called ZGaming \cite{ZGaming} shows that video prediction is a feasible option to enable zero latency cloud gaming experiences by predicting synthetic foreground objects in the virtual world. We consider that video and haptic prediction are the keys to enabling zero latency interaction among remote locations by compensating for network propagation delays.  \looseness=-1

Early video prediction techniques, such as PredNet \cite{PredNet} and CrevNet \cite{CrevNet}, introduce state models to predict the unknown future state. 
However, recently, video prediction models, such as CorrWise \cite{CorrWise} and DMVFN \cite{DMVFN}, have switched to a stateless model. Furthermore, there are approaches that utilize extra information such as semantic segmentation maps in addition to RGB frames \cite{Seg2vid,Vid2vid}. 

In this paper, we propose a simple neural network for video prediction based on the stateless model, called IFRVP, which is an extension of IFRNet \cite{IFRNet}, that can be inferred in real-time.
We first propose three model training methods that extend a video interpolation model to handle video prediction tasks and compare them from multiple viewpoints including prediction accuracy, memory (VRAM usage), and computational cost. 
Then, we show that the recursive model suffers from error accumulation when predicting distant frames, while other models are immunized and highly preferable in zero-latency applications.
After that, we introduce an ELAN-based residual block \cite{ELAN}, which is commonly used in object detection networks, to improve the performance and precision of our prediction network.

Our main contributions are listed below: 
\begin{itemize}
\item We propose three new methods to train video frame interpolation models for video prediction as follows:
\begin{enumerate}
    \item \textbf{Recurrent Prediction}, which uses the two latest frames and recursively predicts the next frame.
    \item \textbf{Arbitrary Prediction}, which enables prediction of any timestep in a single inference by timestep embedding and yields faster inference speed and higher accuracy at large timesteps.
    \item \textbf{Independent Prediction}, which utilizes multiple models corresponding with possible timesteps. This enables prediction of any timestep in a single inference and achieves the best accuracy of the three methods due to larger model capacity.
\end{enumerate}

\item We propose an ELAN-based residual block, which is lightweight and efficient, and implemented in the IFRVP model.
\item We achieved a state-of-the-art speed-accuracy trade-off, offering 20\% faster inference speed, while yielding a higher MS-SSIM than DMVFN.

\end{itemize}

\section{Background}

\subsection{Video Prediction}
Video prediction has been the main interest of many researchers during the last ten years \cite{survey1}. 
Possible applications include collision avoidance in autonomous driving, which enhances operational safety. 
As mentioned earlier, early video prediction models utilize the model architecture with internal states like LSTM to exploit spatio-temporal relations, such as PredNet \cite{PredNet} and CrevNet \cite{CrevNet}. 
However, this approach is computationally intensive and also requires pre-conditioning frames as the input. 
To alleviate this problem, simplified and stateless video prediction models have been proposed, such as DVF \cite{DVF}, OPT \cite{OPT}, CorrWise \cite{CorrWise}, and DMVFN \cite{DMVFN}. Instead of relying on previous states, these model takes advantage of optical flows to handle the prediction tasks.

\subsection{Frame Interpolation and Extrapolation}

Frame interpolation can be classified into optical-flow-based methods and kernel-based methods \cite{survey2}. 
In optical-flow-base methods, the frame interpolation can be formulated as follows. 
We denote reference RGB frames by \begin{math}I_{0}\end{math} and \begin{math}I_{1}\end{math}, and target timestep by \begin{math}t\end{math}, where \begin{math}0 \le t \le 1\end{math}. 
After that, We denote backward warping by $\overleftarrow{\mathcal{W}}$, optical flows by \begin{math}F_{t\rightarrow0}\end{math} and \begin{math}F_{t\rightarrow1}\end{math}, and a pixel weight map by \begin{math}M\end{math}, where \begin{math}0 \le M \le 1\end{math}. 
Hence, the interpolated frame \begin{math}\hat{I}_{t}\end{math} is given by:

\begin{equation*}
{\hat{I}}_t = M \odot {\hat{I}}_{t\leftarrow0} + (1-M) \odot {\hat{I}}_{t\leftarrow1}
\end{equation*}
\begin{equation*}
{\hat{I}}_{t\leftarrow0} = \overleftarrow{\mathcal{W}}(I_0, F_{t\rightarrow0}), \qquad {\hat{I}}_{t\leftarrow1} = \overleftarrow{\mathcal{W}}(I_1, F_{t\rightarrow1})
\end{equation*}
\noindent
where $M$ is generally proportional to the distance to the reference frames. 

Recent works utilize neural networks to improve the prediction accuracy of the optical flow $F$. 
RIFE \cite{RIFE} applies a coarse-to-fine network to predict the optical flow, interpolates pixels, and refines the interpolated pixels by the additional network. 
IFRNet \cite{IFRNet} extends RIFE by unifying the optical flow estimation and the pixel refinements into a single network, significantly improving the performance. 
SLAMP \cite{SLAMP} also employs the optical flow for video prediction, but one of the warped images is replaced by the appearance prediction that is effective in static regions.  
SLAMP also utilizes LSTM to manage temporal probability models, increasing model complexity as mentioned previously. 
Voxel flow is an extension of the optical flow that represents 3D spatio-temporal relations of pixels, which is also utilized in frame interpolation such as DVF \cite{DVF} and DMVFN \cite{DMVFN}. 

Extension of frame interpolation to frame extrapolation can be carried out in a straightforward manner. 
Considering the prediction of the next frame $I_{t+1}$ from the reference frames $I_{t-1}$ and $I_t$. 
Let \begin{math}F_{t+1\rightarrow t}\end{math} and \begin{math}F_{t+1\rightarrow t-1}\end{math} represent estimated optical flows or voxel flows, then the frame extrapolation is formulated by the following equations: 
\begin{equation*}
{\hat{I}}_{t+1} = M \odot {\hat{I}}_{t+1\leftarrow t} + (1-M) \odot {\hat{I}}_{t+1 \leftarrow t-1}
\end{equation*}
\begin{equation*}
{\hat{I}}_{t+1\leftarrow t} = \overleftarrow{\mathcal{W}}(I_t, F_{t+1\rightarrow t})
\end{equation*}
\begin{equation*}
{\hat{I}}_{t+1 \leftarrow t-1} = \overleftarrow{\mathcal{W}}(I_{t-1}, F_{t+1 \rightarrow t-1})
\end{equation*}

\subsection{Zero-Latency Interaction by Video Prediction}

\begin{figure}[t!]
\centering\includegraphics[width=\linewidth]{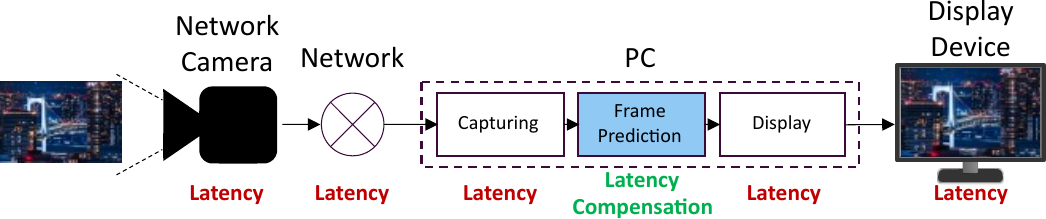}
\vspace{-5mm}
\caption{Latency compensation by video prediction.}
\label{fig:LatencyCompensation}
\vspace{-5.5mm}
\end{figure}

\begin{figure*}[t!]
\centering
\begin{subfigure}{0.1\linewidth}
    \centering\includegraphics[width=0.8\linewidth]{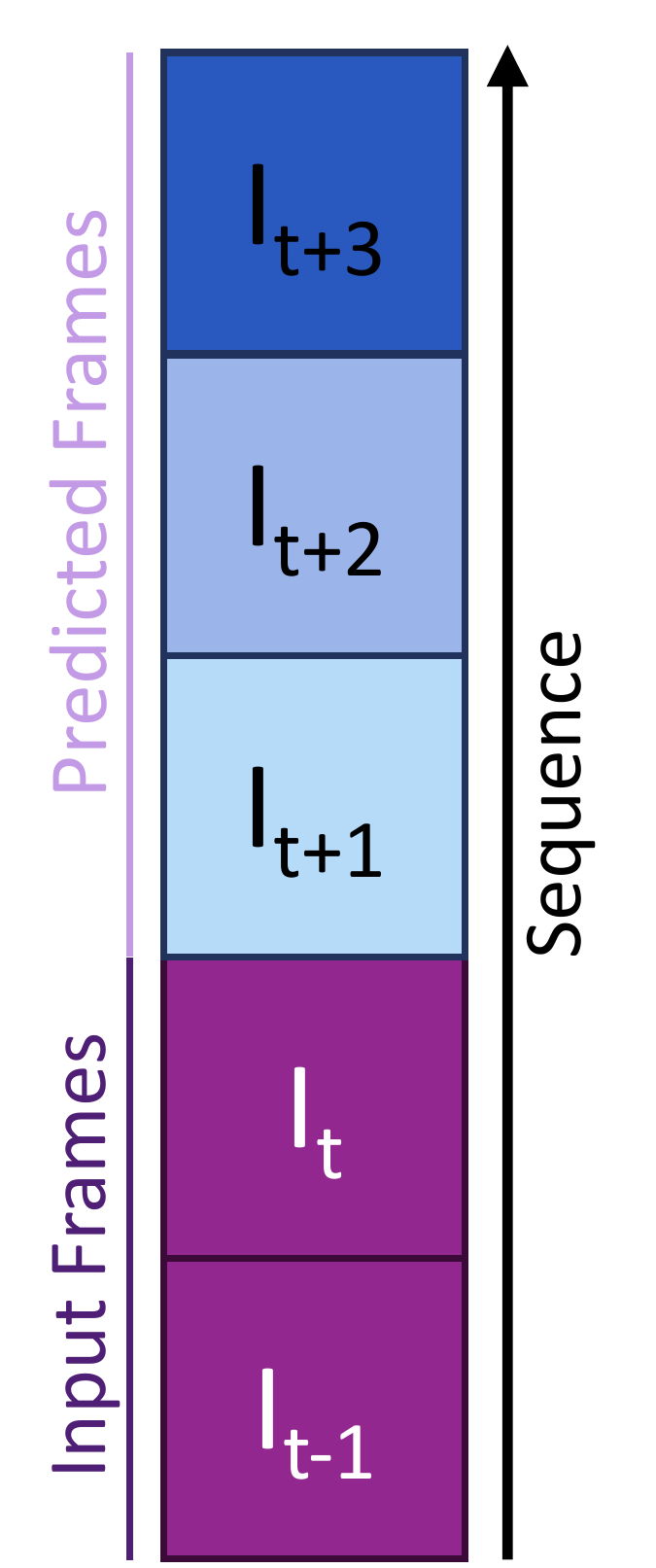}
\end{subfigure}
\begin{subfigure}{0.25\linewidth}
  \centering\includegraphics[width=0.85\linewidth]{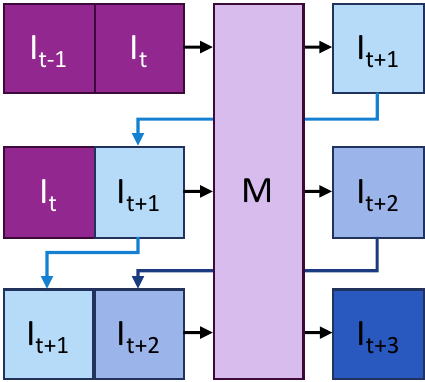}
  
  \caption{Recurrent Prediction}
  \label{fig:RecurrentPredictionTraining}
  \vspace{-3.5mm}
\end{subfigure}
\begin{subfigure}{0.25\linewidth}
  \centering\includegraphics[width=0.85\linewidth]{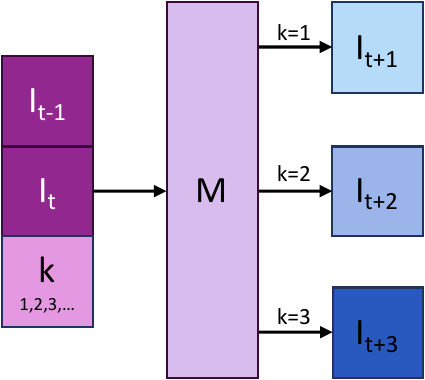}
  \caption{Arbitrary Prediction}
  \label{fig:ArbitraryPredictionTraining}
  \vspace{-3.5mm}
\end{subfigure}
\begin{subfigure}{0.25\linewidth}
  \centering\includegraphics[width=0.85\linewidth]{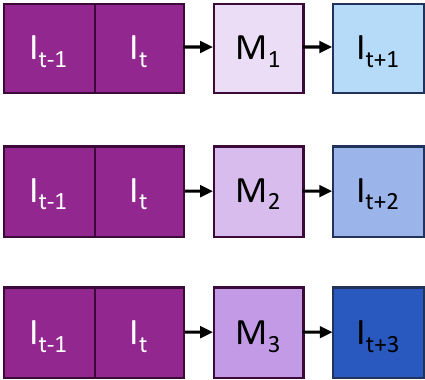}
  \caption{Independent Prediction}
  \label{fig:IndependentPredictionTraining}
  \vspace{-3.5mm}
\end{subfigure}

\caption{Illustrations of the three proposed training methods.}
\label{fig:TrainingMethod}
\vspace{-5.5mm}
\end{figure*}

Latency is inevitable when transmitting images via offsite remote access or long-distance surveillance systems. While acceptable in some cases, latency is one of the major issues in real-time delay-sensitive mission-critical systems, such as remote operation and remote-controlled vehicles with haptic devices, where low-latency video feed is critical for operational safety. 
In these cases, network delay is usually the main cause of the latency and jitter, resulting in fluctuations in total system latency. 
Moreover, each device in the system also introduces extra processing delay due to the nature of its operation. 
By predicting future video frames based on known data, video prediction can be utilized to compensate for the total system delay as seen in \cref{fig:LatencyCompensation}, reducing the latency perceived by the user to almost zero \cite{Teleoperation,Compensation,5GNR,Zero}.
While the figure shows the latency compensation mechanism being implemented on the receiver side, it can also be done on the transmitter side or cloud systems depending on scenarios. \looseness=-1

\vspace{-1mm}
\subsection{Efficient Layer Aggregation Network (ELAN)}
Designing a high-efficiency and high-quality expressive network architecture has always been vital, especially for image classification and object detection tasks on mobile and edge devices. Among many studies on efficient network architectures, the Efficient Layer Aggregation Network (ELAN) \cite{ELAN}, adopted by YOLOv7 \cite{yolov7}, has recently gained a lot of attention in object detection research due to its efficiency. \looseness=-1

\vspace{-1mm}
\section{Proposed Method}

\subsection{Training Extrapolation Models for Video Prediction}

Our main objective is to extend video interpolation models to tackle video prediction tasks. While video interpolation infers $\hat{I}_t$ using ${I}_{t-1}$ and ${I}_{t+1}$, video prediction focuses on inferring $\hat{I}_{t+1}$ using ${I}_{t-1}$ and ${I}_{t}$. Hence, there is a similarity between video interpolation and video prediction tasks. By rearranging the order of the frames in the training dataset, video interpolation models can be easily extended to handle video prediction tasks. We propose three training methods to train video interpolation models for video prediction tasks which only need minor modifications in the model architecture and training code. Each method presents a different approach to predict future timesteps, as shown in \cref{fig:TrainingMethod}. 

\vspace{-1mm}
\subsubsection{Recurrent Prediction Training}

This approach is similar to existing works such as DMVFN \cite{DMVFN}, DVF \cite{DVF}, and PredNet \cite{PredNet}. As illustrated in \cref{fig:RecurrentPredictionTraining}, model are trained to predict an unseen frame ${\hat{I}_{t+1}}$ using $I_{t}$ and $I_{t-1}$. During inference, recurrent prediction can be represented by the following formula, where $f$ is the video prediction model:

\begin{equation*}
{\hat{I}}_{t+k} = \underbrace{(f\circ f\circ f\circ \cdots \circ f)}_{\times k}({I}_{t},{I}_{t-1})
\end{equation*}
The main advantage of this method is the simplicity of implementation. Typically, training of video interpolation models involves loading three consecutive video frames. Then, the model is trained to predict the intermediate (second) frame based on the input of the first and the third frames. By swapping the second and the third frames, the model can be repurposed for video prediction. Because this method only requires rearranging the triplets in datasets, it is widely applicable to most of the existing interpolation models. 

However, recurrent prediction only predicts one timestep at a time, in which the prediction of ${\hat{I}_{t+k}}$ requires $\hat{I}_{t+k-1}$ and $\hat{I}_{t+k-2}$. If the prediction of ${\hat{I}_{t+k}}$ is desired, but the input data only contain $I_{t}$ and $I_{t-1}$, then inference has to be repeated $k$ times, resulting in a linearly increased computational cost proportional to $k$. Furthermore, since the subsequent prediction relies on the past output, the prediction error is propagated to the following frames. Therefore, prediction errors are significantly accumulated for the large timestep $k$.

\vspace{-1mm}
\subsubsection{Arbitrary Prediction Training}

In order to reduce the computational cost and avoid error propagation as with the recurrent prediction, we propose arbitrary prediction training, shown in \cref{fig:ArbitraryPredictionTraining}. Arbitrary timestep interpolation infers the intermediate frame $\hat{I}_{t+k}$ at an arbitrary \textit{real} timestep $k \in (-1,1)$ from two frame: $I_{t-1}$ and $I_{t+1}$ \cite{IFRNet}. Inspired by this, we utilize the previous two frames $I_t$ and $I_{t-1}$ to predict an arbitrary future frame $\hat{I}_{t+k}$ at a \textit{discrete} timestep $k > 0$, which can be represented as follows:
\begin{equation*}
    {\hat{I}_{t+k}} = f(I_{t},I_{t-1},k), k \in \mathbb{Z}^{+}
\end{equation*}

As this method performs all predictions in a single step, it is computationally efficient even for a large $k$, and highly applicable to realize zero-latency transmission. Additionally, this method is highly flexible as it is able to predict a future frame at any given timestep. This allows it to instantaneously offset the network jitter and debounce the output; we simply change the prediction timestep $k$ to cope with the fluctuating round trip time. This ultimately enables adaptive latency compensation even in unstable network conditions. Moreover, unlike the recurrent prediction, this method does not propagate errors to the subsequent timestep, as it only refers to the actual frames $I_{t}$ and $I_{t-1}$.

However, this method requires timestep to be embedded into the model, although many video interpolation models already have such functionality. Additionally, sufficient learning of timestep embedding is challenging and requires long sequences of consecutive video frames. Lastly, larger models are necessary for this method to be effective as small models do not have enough capacity to learn arbitrary timesteps.

\vspace{-1.5mm}
\subsubsection{Independent Prediction Training}

To achieve high-accuracy single-inference video prediction for any arbitrary timesteps, we further propose independent prediction training, shown in \cref{fig:IndependentPredictionTraining}. In this method, we train $n$ models, each optimized for the inference of a specific timestep. In other words, given $1 \leq k \leq n$, model $f_k$ is trained to specifically predict $\hat{I}_k$, which can be formulated as:

\begin{equation*}
  {\hat{I}_{t+k}} = f_k(I_{t},I_{t-1}),\, n \in \mathbb{Z}^+,\, k \in [1, n] \cap \mathbb{Z}^+
\end{equation*}
Compared to arbitrary prediction, independent prediction training eliminates the time embedding requirement, which enables this method to be applied to most video interpolation models, while still able to achieve one-step inference. Due to the larger model capacity, which scales proportionally to $n$, the prediction accuracy is superior to the two former methods. 

However, the training expenses and the total model size also grow linearly with $n$. The practical limitation of employing multiple models is that in a high-jitter network environment, multiple models each trained to predict different timestep $k$ have to be loaded into the memory to cope with varying latency, resulting in a higher memory demand.

\begin{table}[t]
  \caption{Comparison of the proposed training methods.}
  \label{tab:Comparison}
  \centering
  \setstretch{0.85}
  \rowcolors{1}{}{gray90}
  \resizebox{8.4cm}{!}{
    \renewcommand{\arraystretch}{1.3}
        \begin{tabular}{c c c c}
        \toprule
        & Recurrent & Arbitrary & Independent\\ \midrule
        Accuracy for Small k & $\bigstar\bigstar\bigstar$ & $\bigstar\bigstar\openbigstar$  & $\bigstar\bigstar\bigstar$  \\ 
        Accuracy for Large k & $\bigstar\openbigstar\openbigstar$& $\bigstar\bigstar\openbigstar$  & $\bigstar\bigstar\bigstar$ \\ 
        1-step Inference     & \xmark  & \cmark  & \cmark \\ 
        VRAM Friendly        & \cmark  & \cmark  & \xmark \\ 
        \bottomrule
    \end{tabular}
  }
  \vspace{-4.5mm}
\end{table} 

\vspace{-1.5mm}
\subsubsection{Comparison of Training Models}
The comparison of all three proposed methods is shown in \cref{tab:Comparison}. 
As seen from the table, while the IFRVP model trained with recurrent prediction yields good prediction accuracy for small $k$, the model suffers from error propagation, resulting in poor accuracy as $k$ increases. On the other hand, the model trained with arbitrary training performed noticeably better, especially for the large $k$, demonstrating that the model is capable of learning various embedded $k$. However, the model performs slightly worse than the former method at small timestep $k$ likely due to the smaller model capacity. This issue is ultimately addressed via independent prediction training, where multiple models are each trained for prediction at a specific timestep, resulting in a much higher model capacity, but at a cost of higher memory consumption.

Then, we further demonstrate the effectiveness of the above methods by extending the baseline video frame interpolation model, the IFRNet \cite{IFRNet}, to perform video prediction tasks. The IFRNet is selected due to its architectural simplicity and efficiency, and its unification of optical flow estimation and pixel refinements network. Additionally, the model already features timestep embedding, which allows the implementation of the arbitrary method. We call our derived video prediction models \textbf{IFRVP}. \looseness=-1

\vspace{-2mm}
\subsection{ELAN-based Residual Block for IFRVP}

\begin{figure}[t!]
\centering
\begin{subfigure}{0.32\linewidth}
 \centering\includegraphics[width=0.98\linewidth]{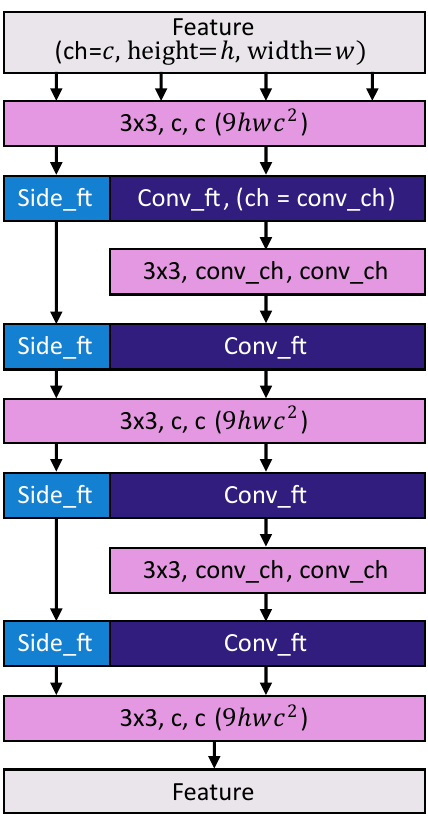}
  \caption{IFRNet \cite{IFRNet}}
  \label{fig:OriginalBlock}
  \vspace{-3mm}
\end{subfigure}
\begin{subfigure}{0.32\linewidth}
  \centering\includegraphics[width=0.98\linewidth]{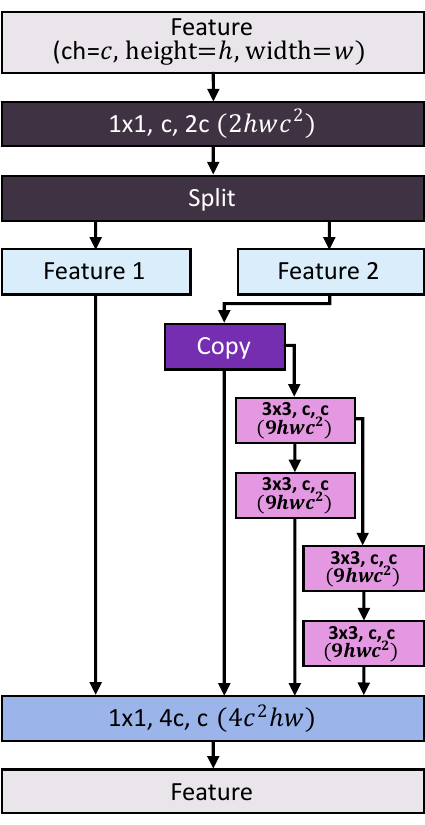}
  \caption{ELAN \cite{ELAN}}
  \label{fig:ELANBlock}
  \vspace{-3mm}
\end{subfigure}
\begin{subfigure}{0.32\linewidth}
  \centering\includegraphics[width=0.98\linewidth]{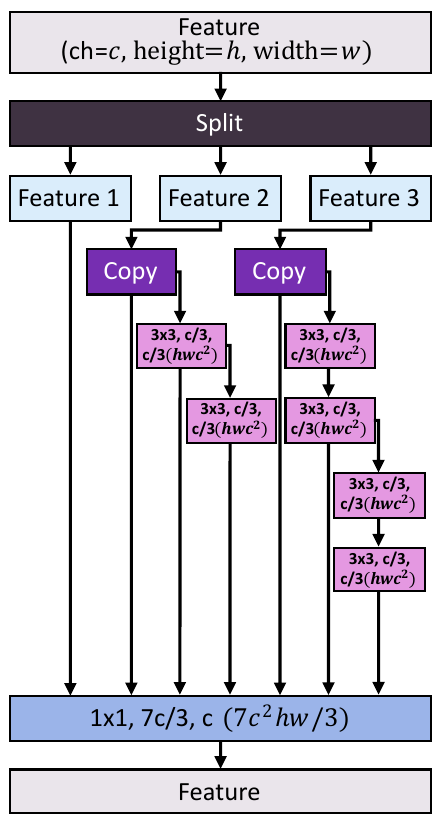}
  \caption{Proposed}
  \label{fig:ProposedBlock}
  \vspace{-3mm}
\end{subfigure}

\caption{Architecture comparisons of various residual blocks.
Estimated FLOPs is written in convolution block}
\vspace{-5mm}
\label{fig:ResidualBlock}
\end{figure}

\begin{table*}[t]
  \caption{Quantitative comparison of different methods on the Cityscapes dataset \cite{cityscapes}.}
  \setstretch{0.94}
  \label{tab:scorecomparison}
  \centering
    \renewcommand{\arraystretch}{1.1}
      \begin{tabular}{c c c c c c}
        \toprule
         & \multirow{3}{*}{\shortstack{GFLOPs}} &\multirow{3}{*}{\shortstack{method}}& \multicolumn{3}{c}{MS-SSIM $\uparrow$}\\\cmidrule(lr){4-6} 
        & && t+1& t+3&t+5\\ 
        
        \midrule
        PredNet \cite{PredNet}& 62.62&recurrent
    & 0.8403& 0.7925&0.7521\\ \hline
    DVF \cite{DVF}
    & 409.78&recurrent
    & 0.8385& 0.7623&0.7111\\ \hline
    CorrWise \cite{CorrWise}
    & 944.29&recurrent
    & 0.9280& N/A&0.8390\\ \hline
    OPT \cite{OPT}
    & 313482.15 &recurrent
    & 0.9454 & 0.8689 &0.8040  \\ \hline
        DMVFN \cite{DMVFN}
    & 12.71
    &recurrent
    & 0.9573& 0.8924&0.8345\\ \hline

    \multirow{3}{*}{\shortstack{IFRVP}}& \multirow{3}{*}{57.71}
    & recurrent
    &  \textbf{0.9773}& 0.8338&0.7206
\\ \cline{3-6}

    & 
    & arbitrary
    &  0.9577& 0.9009&0.8538
\\ \cline{3-6}
     
    & 
    & independent
    &\textbf{0.9773}& \textbf{0.9214}& \textbf{0.8710}\\ \hline

    \multirow{3}{*}{\shortstack{IFRVP-Fast}}
    & \multirow{3}{*}{\textbf{9.9}} & recurrent
    & \textbf{\textcolor{blue}{0.9625}}& 0.8858& 0.7909
\\\cline{3-6}

    & 
    & arbitrary
    & 0.9307& 0.8689& 0.8184
\\\cline{3-6}
     
    & 
    & independent
    &\textbf{\textcolor{blue}{0.9625}}& \textbf{\textcolor{blue}{0.8946}}& \textbf{\textcolor{blue}{0.8438}}
\\\hline
\end{tabular}
\vspace{-5mm}
\end{table*} 

We found that IFRVP's computational complexity has the potential to be further reduced. Therefore, we propose a lightweight residual block inspired by ELAN \cite{ELAN}. Then, we demonstrate its effectiveness by replacing the residual blocks in IFRVP decoder with our proposed one. We name this derived model \textbf{IFRVP-Fast}. The architecture comparison between the original residual blocks and our proposed residual block is illustrated in \cref{fig:ResidualBlock}.

The original ELAN first doubles the number of channels using point-wise convolution, then partially applies 3×3 convolution. This architecture enhances the learning capability by increasing the variety of the gradient paths.
Our proposed block does not double the number of channels but simply splits the input feature to reduce the total computational cost. 
The computational complexity of the original ELAN block can be calculated as $42c^2hw$ FLOPs, where $c,h$ and $w$ are the number of channels, height, and width of the input feature, respectively. On the other hand, our proposed block has the computational complexity of $25c^2hw/3$ FLOPs, a 5× reduction over the original ELAN block. On the other hand, the IFRNet residual block has a computational cost of at least $27c^2hw$ FLOPs. Hence, our method yields a 3× computation reduction over the IFRNet residual block, which enables our proposed block to be double-stacked, increasing the depth of the network to enhance its prediction accuracy while keeping the computational complexity below the original methods.

With the implementation of our newly proposed residual block and further feature channel reduction, IFRVP-Fast is a very lightweight model with a computational cost of less than 10 GFLOPs, a significant improvement over the state-of-the-art video prediction model DMVFN \cite{DMVFN}.

\vspace{-3mm}
\section{EXPERIMENTS}

\vspace{-1mm}
\subsection{Experiment Setup}

We trained our models with the Cityscapes \cite{cityscapes} dataset, which contains urban street scenes from multiple cities, providing a good representation of the actual video feed during autonomous driving. Triplet sequences are constructed for recurrent and independent prediction, while septuplet sequences are built for the arbitrary approach, which makes our results compatible with the result reported in DMVFN \cite{DMVFN}. We calculate the prediction accuracy from $t+1$ to $t+5$ with its corresponding $k$ of $\{1,2,3,4,5\}$ for arbitrary prediction and independent prediction training.

Overall, there are six variations of our proposed models. Two models, \textbf{IFRVP} and \textbf{IFRVP-Fast}, are trained with three different prediction methods: \textbf{recurrent}, \textbf{arbitrary}, and \textbf{independent} approaches. We train the recurrent and the independent prediction models for 200 epochs with a batch size of 12. However, due to septuplet sequences, the arbitrary method predicts five frames using two reference frames in each training step. Hence, to keep the training cost roughly equal among the three approaches, we reduced the batch size to 3 and only trained the arbitrary model for 100 epochs. Our models were optimized to minimize the L1 loss, which is calculated based on the Laplacian pyramid representations \cite{laplacian} extracted from each pair of images. Furthermore, we also evaluated and compared our methods against DMVFN \cite{DMVFN}, the state-of-the-art video prediction model, as well as other existing methods using the same dataset. \looseness=-1 
\vspace{-1mm}
\subsection{Quantitative Evaluation}

MS-SSIM \cite{msssim} is used as the evaluation metrics. Additionally, we also calculate the computational complexity (FLOPs) of each model. Because DMVFN is evaluated using the downscaled Cityscapes frames in the original work, we followed their configuration by downscaling the images from 2048$\times$1024 to 1024$\times$512. 

\begin{figure}[t!]
\centering
\includegraphics[width=8.5cm]{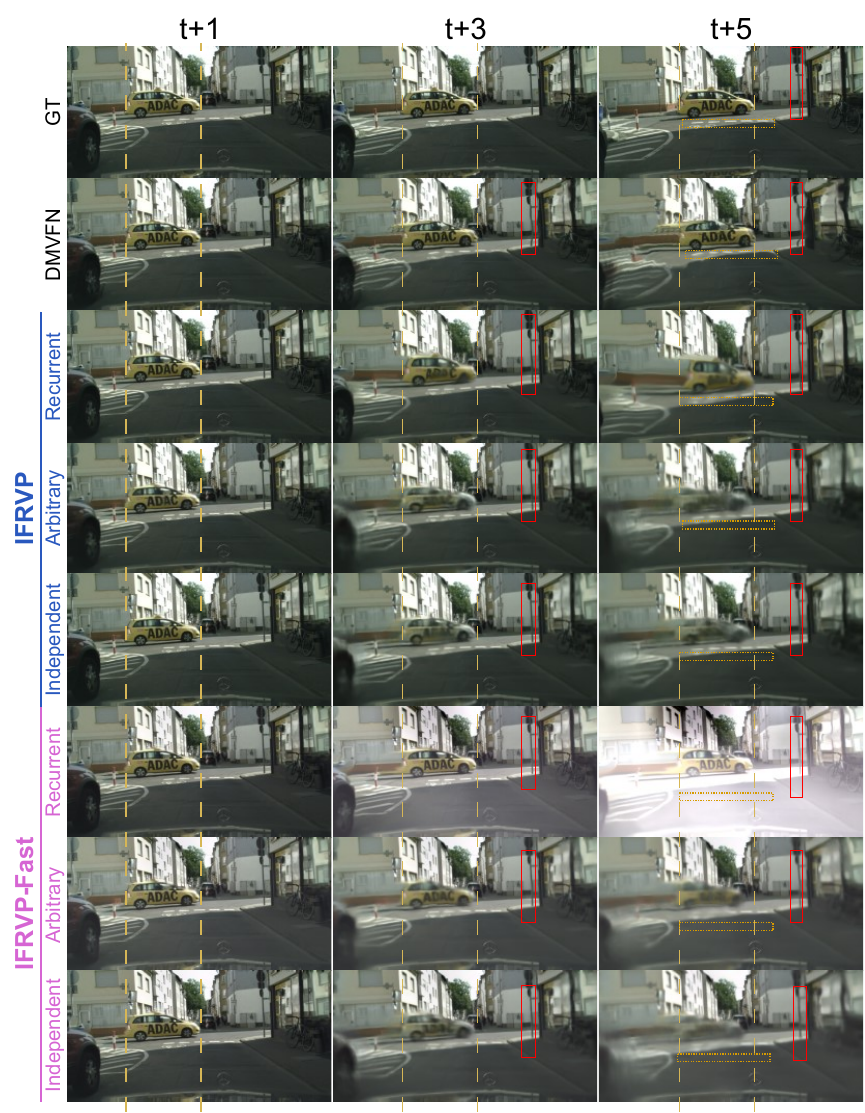}
\vspace{-3.5mm}
  \caption{Qualitative prediction comparison on Cityscapes \cite{cityscapes}.}
  \label{fig:Qualitativepredictioncomparison}
  \vspace{-6mm}
\end{figure}

As shown in the \cref{tab:scorecomparison}, IFRVP achieves the best MS-SSIM among all the methods. Particularly, the IFRVP trained with the independent prediction method achieved state-of-the-art results. Although IFRVP requires around 4.5× higher computational cost compared to DMVFN, this cost remains constant regardless of $k$ as the prediction is being done in a single step. On the other hand, the computational cost of DMVFN scales proportionally to $k$. Thus, the DMVFN model will be more computationally expensive when $k>4$. For IFRVP trained with the arbitrary method, higher MS-SSIM is also achieved when compared to DMVFN across all timesteps, demonstrating that IFRVP is highly applicable for latency compensation in high-jitter network environments. IFRVP-Fast also outperforms DMVFN in the MS-SSIM metric requiring 20\% lower computational cost, yielding the best speed-accuracy trade-off. 
\looseness=-1
\vspace{-1mm}
\subsection{Qualitative Evaluation}

For the qualitative comparison, we compare the prediction results of Cityscapes test dataset using IFRVP. The results of the qualitative comparison are shown in \cref{fig:Qualitativepredictioncomparison}. The red box highlights the distortion exhibited by each model on the traffic sign. First of all, recurrent prediction is subjectively sharper than arbitrary prediction for $\hat{I}_{t+3}$ as the timestep $k$ is small, which is coherent with the results in \cref{tab:Comparison}. Furthermore, for the predicted image at $\hat{I}_{t+5}$, we find that while DMVFN is quite capable of maintaining the object shape, the contours are considerably distorted. However, our arbitrary and independent methods keep the object shape mostly intact, with only a slight blur being observed. The only exception is the recurrent predicted image, which suffered from propagated distortions, and the length of the yellow car was shortened. 
Although IFRVP-Fast with recurrent prediction fails to predict brightness correctly, the shape of objects is still kept compared to DMVFN despite having more blur.

\looseness=-1
\vspace{-1mm}
\subsection{Demonstration Video}

\begin{figure}[t]
\centering
\includegraphics[width=8.5cm]{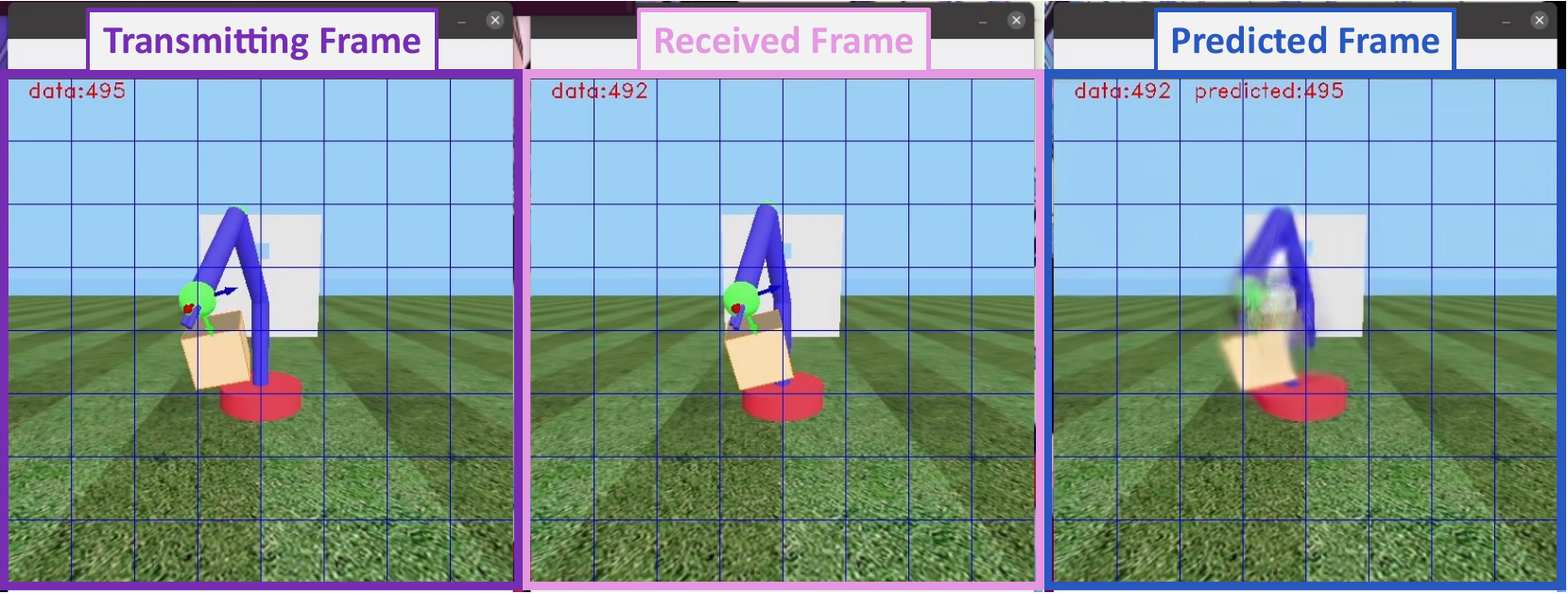}
\vspace{-2mm}
  \caption{Screenshot of demo video. To simulate network latency, we artificially added a 100 ms delay (equivalent to 3 frames; 30 fps video was used). The predicted frame was based on the two latest received frame.}
  \label{fig:ScreenshotOfDemoMovie}
  \vspace{-5mm}
\end{figure}
We created a demonstration video to showcase our models performing real-time inferences to achieve zero latency similar to the one in \cite{LICoris}. To prove that IFRVP can run on edge devices, we used an affordable GPU (NVIDIA GeForce RTX 3070) for the demonstration. In the video, for our first demonstration, the videos from Xiph.org \cite{xiph} with the resolution of 352$\times$288 were used. Then, the simulation videos of a user operating a robot arm, which has a resolution of 512$\times$512, were used for the subsequent demonstrations. The demo video can be accessed at \textbf{http://bit.ly/IFRVPDemo}, and the screenshots can be seen in \cref{fig:ScreenshotOfDemoMovie}.

As shown in the figure, the predicted video frame is quite similar to the actual frame on the transmitting side, which shows that our proposed model can be effectively used to compensate the network latency in video communication applications. 
Furthermore, our proposed network empirically runs at 130 and 70 frames per second at the resolution of 352$\times$288 and 512$\times$512, respectively.
\looseness=-1
\vspace{-1.5mm}
\section{Conclusions}
In this paper, we proposed three methods to extend video interpolation models for video prediction tasks, which can be applied to realize zero-latency telecommunication. We implemented IFRVP \cite{IFRNet}, a lightweight video interpolation model, and then applied the training methods to the model. Secondly, we propose a lightweight residual block based on ELAN, which significantly reduces computational costs while retaining high prediction accuracy. We then applied the residual block to our proposed IFRVP-Fast, which has a computational complexity of under 10 GFLOPs. From our evaluations, both IFRVP and IFRVP-Fast yielded state-of-the-art video prediction performance, while also substantially better at preserving object edges compared to the baselines.
For future work, we will continue to pursue more efficient and lightweight models for video prediction to further realize a reliable zero-latency interaction over communication networks. \looseness=-1

\vspace{-1.5mm}
\section{Acknowledgement}
This research was supported by the commissioned research (No. JPJ012368C03801) by National Institute of Information and Communications Technology (NICT), Japan and JSPS KAKENHI Grant Number JP23K16861. Additionally, the first author would like to express his gratitude to Mitsuki Nakae for warm encouragement. 

\setstretch{0.97}
\newcommand{\BIBdecl}{\setlength{\itemsep}{0.1 em}}


\end{document}